%% file: main.tex
\documentclass[sigconf, nonacm]{acmart}

\newcommand\vldbdoi{XX.XX/XXX.XX}

\newcommand\vldbavailabilityurl{https://github.com/cs-anonymous/eXpath}

\usepackage{algorithmic}
\usepackage{array}
\usepackage{multirow}
\usepackage{makecell}
\usepackage{algorithm}
\usepackage{graphicx} 

\begin{document}
\title{eXpath: Explaining Knowledge Graph Link Prediction with Ontological Closed Path Rules}

\author{Ye Sun}
\orcid{0000-0002-3679-483X}
\affiliation{%
  \institution{School of Computer Science, Beihang University}
  \city{Beijing}
  \country{China}
}
\email{sunie@buaa.edu.cn}

\author{Lei Shi$^\ast$} 
\orcid{0000-0002-1965-2602}
\affiliation{%
  \institution{School of Computer Science, Beihang University}
  \city{Beijing}
  \country{China}
}
\email{leishi@buaa.edu.cn}

\author{Yongxin Tong}
\orcid{0000-0002-5598-0312}
\affiliation{%
  \institution{School of Computer Science, Beihang University}
  \city{Beijing}
  \country{China}
}
\email{yxtong@buaa.edu.cn}

\thanks{$^\ast$Corresponding author.}


\begin{abstract}
Link prediction (LP) is crucial for Knowledge Graphs (KG) completion but commonly suffers from interpretability issues. While several methods have been proposed to explain embedding-based LP models, they are generally limited to local explanations on KG and are deficient in providing human interpretable semantics. Based on real-world observations of the characteristics of KGs from multiple domains, we propose to explain LP models in KG with path-based explanations. An integrated framework, namely eXpath, is introduced which incorporates the concept of relation path with ontological closed path rules to enhance both the efficiency and effectiveness of LP interpretation. Notably, the eXpath explanations can be fused with other single-link explanation approaches to achieve a better overall solution. Extensive experiments across benchmark datasets and LP models demonstrate that introducing eXpath can boost the quality of resulting explanations by about 20\% on two key metrics and reduce the required explanation time by 61.4\%, in comparison to the best existing method. Case studies further highlight eXpath's ability to provide more semantically meaningful explanations through path-based evidence.
\end{abstract}

\maketitle


\vspace{-0.3cm}
\ifdefempty{\vldbavailabilityurl}{}{
\vspace{.3cm}
\begingroup\small\noindent\raggedright\textbf{PVLDB Artifact Availability:}\\
The source code, data, and/or other artifacts have been made available at \url{\vldbavailabilityurl}.
\endgroup
}

\input{tex/1_intro}
\input{tex/2_related}
\input{tex/3_background}

\input{tex/4_framework}

\input{tex/5_experiment}

\section{Conclusion}

In this work, we introduce eXpath, a novel path-based explanation framework designed to enhance the interpretability of LP tasks on KG. By leveraging ontological closed path rules, eXpath provides semantically rich explanations that address challenges such as scalability and relevancy of path evaluation on embedding-based KGLP models. Extensive experiments on benchmark datasets and mainstream KG models demonstrate that eXpath outperforms the best existing method by 12.4\% on $\delta MRR$ in terms of the most important multi-fact explanations. A higher improvement of 20.2\% is achieved when eXpath is further combined with existing methods. Ablation studies validate that the CP rule in our framework plays a central role in the explanation quality, with its removal leading to a 20.3\% average drop in performance.

While our method currently utilizes a small subset of key ontological rules, other rule types, such as unary rules with dangling atoms, are found to have less impact on LP results. This suggests that broader language biases may not always align with the strengths of embedding-based models. Future work can explore the potential of general rule learning on KG and adapt them to the eXpath's overall framework. Additionally, the semantically rich explanations supported by eXpath can benefit from interactive visualization tools, offering enhanced accessibility and understanding of the explanations for both KG experts and non-expert users.



\bibliographystyle{ACM-Reference-Format}
\bibliography{sample}

\end{document}

%% file: tex/1_intro.tex
\section{Introduction}

Knowledge graphs (KGs) \cite{auer2007dbpedia,bollacker2008freebase,mahdisoltani2014yago3} commonly suffer from incompleteness, such that link prediction (LP) becomes a crucial task for KG completion, aiming to predict potential missing relationships between entities within a KG. In the deep learning era, advanced KG embedding models (KGE) such as ComplEx~\cite{trouillon2016complex}, TransE~\cite{wang2014knowledge}, and ConvE~\cite{dettmers2018convolutional} have been applied to perform the LP task successfully. Yet, due to the inherent black-box nature of deep learning, how to interpret these LP models remains a daunting issue for KG applications. For example, in financial KGs used to make high-stake decisions such as fraud or credit card risk detection, interpretability is required not only for customer engagement purpose \cite{nallakaruppan2024credit}, but also by the latest law enforcement \cite{us2021ai_accountability}.

Various methods have been developed to interpret the behaviour of LP models, e.g., to explain graph neural network (GNN) based predictive tasks~\cite{ying2019gnnexplainer, zhang2023page, chang2024path}, embedding-based models~\cite{bhardwaj2021adversarial, zhang2019data}, and providing subgraph-based explanations~\cite{yuan2021explainability, zhang2022gstarx, zhao2023ke}. On KG, the recently proposed adversarial attack methods~\cite{rossi2022explaining, bhardwaj2021adversarial, pezeshkpour2019investigating} become a major class of approaches for explaining LP results. The adversarial method captures a minimal modification to KG as an optimal explanation if only a maximal negative impact is detected on the target prediction. In particular, Kelpie~\cite{rossi2022explaining} introduces entity mimic and post-training techniques to quantify the model's sensitivity to link removal and addition. Despite the success of LP explanation models on KG, they have key limitations in at least two aspects. First, in most methods, only local explanations related to the head or tail entity of the predicted link are considered without exploring the full KG. Second, the explanations generally focus on maximizing computation-level explainability, e.g., the perturbation to predictive power when adding/removing the potential explanation link. They mostly lack semantic-level explainability, which is extremely important for human understanding.

\begin{figure}[t]
\includegraphics[width=\columnwidth, clip, trim=40pt 20pt 20pt 20pt]{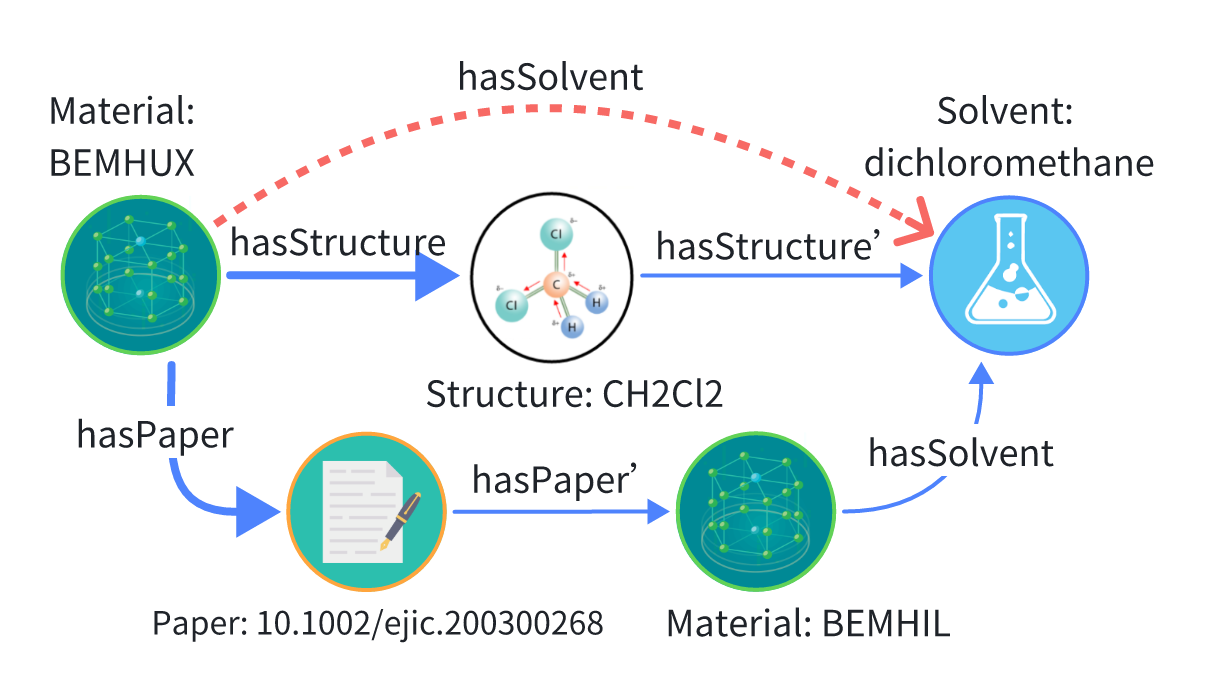}
\vspace{-0.7cm}
\caption{An example of material KG for synthesis route inference. To explain the predicted link ⟨material: BEMHUX, hasSolvent, solvent: dichloromethane⟩ (the dotted red link on the top), two key KG paths (blue links on the middle/bottom) are detected by our method: BEMHUX and dichloromethane sharing the same material sub-structure; BEMHUX appearing in the same paper with another material BEMHIL, which also uses dichloromethane as the solvent. Classical LP explanations (e.g., Kelpie) will select the single-hop links as explanations (thickened blue links).}
\label{MOFexample}
\vspace{-0.5cm}
\end{figure}

In this work, we are motivated by several observations during the real-world deployment of LP models on KG. For instance, in a material knowledge graph of Fig.~\ref{MOFexample}, to explain the fact of a material synthesized within a particular solvent (dotted red link), classical methods only excerpt single-hop links representing certain properties of the material (thickened blue links). In reality, the material expert favours path-based explanations such as the blue paths on the middle/bottom of Fig.~\ref{MOFexample}. The middle path indicates material/solvent sharing the same sub-structure, while the bottom path indicates two materials reported by the same paper/team that potentially share the same solvent environment. These path-based explanations represent fundamental semantics, such as causal relationships with the predicted link. Building on these observations, we propose a path-based explanation framework, namely eXpath, to address the interpretability problem of LP models on KG. Our method not only suggests minimal KG modifications similar to adversarial attack explanations but also highlights semantically meaningful link paths supporting each modification.

Note that the idea of path-based explanation has also been studied in the recent work of Power-Link \cite{chang2024path} and PaGE-Link \cite{zhang2023page}. However, these works focus on explaining GNN-based embedding models and extracting all the potential KG paths up to thousands for a single explanation. In comparison, we consider the explanation of factorization-based embedding models, a mainstream method for KGs. The follow-up adversarial explanation evaluates only a few KG modifications at a time, and it is computationally costly to select the best set of paths from thousands of candidates. Moreover, another pragmatic challenge lies in the evaluation of individual path explanation. While the adversarial method works well in quantifying the effectiveness of a single-link explanation, adding/deleting an entire path can bring more significant change to the KG, hard to evaluate by the same method. The contribution of this work is to address the above challenges as summarized below:

\begin{itemize}
  \item Based on the attributed characteristics of KG, we introduce the concept of relation path, which aggregates individual paths by their relation types. The explanation analysis then works on the level of relation paths, greatly reducing the computational cost while augmenting the semantics of explanations;
  \item On the evaluation of path-based explanations, we propose to borrow ontology theory, particularly the closed path rule and property transition rule, which not only reassures the path-based semantics but also guarantees high-occurrence explanations within the whole KG dataset;
  \item Through extensive experiments across multiple KG datasets and embedding models, we demonstrate the effectiveness of our method, which significantly outperforms existing LP explanation models. Case study also reveals the consistency of path-based explanations with ground-truth semantics.
\end{itemize} 

%% file: tex/2_related.tex
\section{Related Work}

\subsection{The Explanation of Knowledge Graph Link Prediction (KGLP)}

Explainability in Knowledge Graph Link Prediction (KGLP) is a critical area of research due to the increasing complexity of models used in link prediction tasks. General-purpose explainability techniques are widely used to understand the input features most responsible for a prediction. LIME~\cite{ribeiro2016should} creates local, interpretable models by perturbing input features and fitting regression models, while SHAP~\cite{lundberg2017unified} assigns feature importance scores using Shapley values from game theory. ANCHOR~\cite{ribeiro2018anchors} identifies consistent feature sets that ensure reliable predictions across samples. These frameworks have been widely adopted in various domains, including adaptations for graph-based tasks, although their application in link prediction for knowledge graphs remains limited.

GNN-based LP explanation primarily focuses on interpreting the internal workings of graph neural networks for link prediction. Techniques like GNNExplainer~\cite{ying2019gnnexplainer} and PGExplainer~\cite{luo2020parameterized} identify influential subgraphs through mutual information, providing insights into node and graph-level predictions, although they are not directly applicable to link prediction tasks. Other methods, such as SubgraphX~\cite{yuan2021explainability} and GStarX~\cite{zhang2022gstarx}, use game theory values to select subgraphs relevant to link prediction. At the same time, PaGE-Link~\cite{zhang2023page} argues that paths are more interpretable than subgraphs and extends the explanation task to the link prediction problem on heterogeneous graphs. Additionally, Power-Link~\cite{chang2024path}, a path-based KGLP explainer, leverages a graph-powering technique for more memory-efficient and parallelizable explanations. However, GNN-based explainability techniques are limited to GNN-based LP models and do not extend to embedding-based approaches.

\subsection{Adversarial Attacks on KGE}

Adversarial attacks on KGE models have gained attention for assessing and improving their robustness. These attacks focus primarily on providing local, instance-level explanations. The goal is to introduce minimal modifications to a knowledge graph that maximizes the negative impact on the prediction. Methods are typically categorized as white-box or black-box approaches.

White-box methods propose algorithms that approximate the impact of graph modifications on specific predictions and identify crucial changes. Criage~\cite{pezeshkpour2019investigating} applies first-order Taylor approximations for estimating the impact of removing facts on prediction scores. Data Poisoning~\cite{bhardwaj2021adversarial, zhang2019data} manipulates embeddings by perturbing entity vectors to degrade the model's scoring function, highlighting pivotal facts during training. ExamplE~\cite{janik2022explaining} introduce ExamplE heuristics, which generate disconnected triplets as influential examples in latent space. KE-X~\cite{zhao2023ke} leverages information entropy to quantify the importance of explanation candidates and explains KGE-based models by extracting valuable subgraphs through a modified message-passing mechanism. While these white-box methods offer valuable insights, they often require full access to model parameters, making them impractical for real-world applications.

Recent research has also focused on black-box adversarial attacks, which do not require knowledge of the underlying model architecture. KGEAttack~\cite{betz2022adversarial} uses rule learning and abductive reasoning to identify critical triples influencing predictions, offering a model-agnostic alternative to white-box methods. While this study is closely related to ours, it employs simpler rules and does not consider the support provided by multiple or long rules for the facts. Kelpie~\cite{rossi2022explaining} explains KGE-based predictions by identifying influential training facts, utilizing mimic and post-training techniques to sense the underlying embedding mechanism without relying on model structure. However, these methods are limited to fact-based explanations that focus only on local connections to the head or tail entity without capturing the multi-relational context needed for full interpretability.

\subsection{Ontological Rules for Knowledge Graph}

Ontological rules for knowledge graphs have been a prominent area of research, as they provide symbolic and interpretable reasoning over knowledge graph data. AMIE~\cite{galarraga2013amie, galarraga2015fast} and AnyBURL~\cite{meilicke2020reinforced, meilicke2024anytime} extract rules from large RDF knowledge bases and employ efficient pruning techniques to generate high-quality rules, which are then used to infer missing facts in knowledge graphs. Path-based rule learning has also been explored to improve link prediction explainability. Bhowmik~\cite{bhowmik2020explainable} proposes a framework emphasizing reasoning paths to improve link prediction interpretability in evolving knowledge graphs. RLvLR~\cite{omran2018scalable, omran2019embedding} combines embedding techniques with efficient sampling to optimize rule learning for large-scale and streaming KGs. While these methods excel in structural reasoning, they are limited in directly explaining predictions made by embedding-based models, highlighting a gap in integrating rule-based reasoning with KGE interpretability.

Recent works have explored the combination of symbolic reasoning with KGE models. For instance, Guo et al.\cite{guo2018knowledge} introduced rules as background knowledge to enhance the training of embedding models, while Zhang et al.\cite{zhang2019iteratively} proposed an alternating training scheme that incorporates symbolic rules. Meilicke et al.~\cite{meilicke2021naive} demonstrated that symbolic and sub-symbolic models share commonalities, suggesting that KGE models may be explained using rule-based approaches. However, these methods have not been directly applied to explain predictions made by KGE models. While it might be possible to explain a prediction made by a KGE model using a rule-based approach, integrating symbolic reasoning with adversarial attacks remains a challenge.

%% file: tex/3_background.tex
\section{Background and problem definition}
\subsection{KGLP Explanation}

Knowledge Graphs (KGs), denoted as $KG = (\mathcal{E}, \mathcal{R}, \mathcal{G})$, are structured representations of real-world facts, where entities from $\mathcal{E}$ are connected by directed edges in $\mathcal{G}$, each representing semantic relations from $\mathcal{R}$. These edges $\mathcal{G} \subseteq \mathcal{E} \times \mathcal{R} \times \mathcal{E}$, represent facts of the form $f = \langle h, r, t\rangle$, where $h$ is the head entity, $r$ is the relation, and $t$ is the tail entity. Link Prediction (LP) aims to predict missing relations between entities in a KG. The standard approach to LP is embedding-based, where entities and relations are embedded into continuous vector spaces, and a scoring function, $f_r(h,t)$, is used to measure the plausibility of a fact. Evaluation of LP models is typically performed using metrics such as mean reciprocal rank (MRR), which measures how well the model ranks the correct entities when predicting missing heads or tails in the test set $\mathcal{G}_{test}$.

\vspace{-0.2cm}
\begin{equation}
    MRR = \frac{1}{2|\mathcal{G}_{test}|} \sum_{f \in \mathcal{G}_{test}} \left( \frac{1}{\operatorname{rk}_h(f)} + \frac{1}{\operatorname{rk}_t(f)} \right)
\end{equation}
\vspace{-0.2cm}

\noindent where $\operatorname{rk}_t(f)$ represents the rank of the target candidate $t$ in the query $\langle h, r, ?\rangle$, and $\operatorname{rk}_h(f)$ the rank of the target candidate $h$ in the query $\langle ?, r, t\rangle$. 

While embedding-based LP provides accurate predictions, understanding the reasoning behind these predictions is essential for model transparency and trust. To address this, explanation methods for embedding-based LP focus on providing instance-level insights into predictions, revealing underlying features like proximity, shared neighbors, or similar latent factors. However, directly perturbing the model's architecture or embeddings is challenging. As a result, explanation methods often rely on adversarial perturbations within the training data, such as modifications to the neighborhood of the target triple, to assess the robustness of KGE models.

\subsection{Adversarial Attack Problem}

Adversarial attacks in the context of KGLP explanations are designed to assess a model's vulnerability to small changes and evaluate the stability of LP models by intentionally degrading their performance through targeted perturbations in the training data. These attacks provide instance-level adversarial modifications as explanations. Given a prediction \(\langle h, r, t \rangle\), an explanation is defined as the smallest set of training facts that enabled the model to predict either the tail \(t\) in \(\langle h, r, ? \rangle\) or the head \(h\) in \(\langle ?, r, t \rangle\). For example, to explain why the top-ranked tail for \(\langle Barack\_Obama, nationality, ? \rangle\) is 'USA', we identify the smallest set of facts whose removal from the training set \(\mathcal{G}_{\text{train}}\) would cause the model to change its prediction for \(\langle h, r, ? \rangle\) from 'USA' to any entity \(e \neq t\), and for \(\langle ?, r, t \rangle\) from \(h\) to any entity \(e' \neq h\). These facts involve the head and tail entities, as they are crucial to the prediction.

We evaluate the impact of the adversarial attack by comparing standard metrics, such as MRR, before and after the attack. Specifically, we train the model on the original training set and select a small subset of the test set $T \subset \mathcal{G}_e$ as target triples for which the model achieves good predictive performance. After removing the attack set from the training set, we retrain the model and measure the degradation in performance on the target set.

Since we focus on small perturbations, the attack is restricted to deleting a small set of triples. To make this process computationally feasible, we adopt a batch mode where the deletion of one target triple may affect others. If the target sets are small and the predicates contain disjoint entities, dependencies between triples are rare and can typically be neglected. The explanatory capability of the attack is measured by the degradation in MRR, defined as:

\vspace{-0.2cm}
\begin{equation}
\delta MRR(T) = 1 - \frac{MRR_{\text{new}}(T)}{MRR_{\text{original}}(T)}
\end{equation}
\vspace{-0.2cm}

\subsection{Path-Based Exaplanation}

While adversarial attacks focus on identifying critical facts for each prediction, they often lack a clear rationale for why specific facts are considered critical. We observe that certain knowledge graphs, as shown in Fig.~\ref{MOFexample}, exhibit semantically meaningful paths that can enhance the interpretability of predictions.

In this work, we tackle the adversarial attack problem with path-based explanations. Given a prediction $\langle h,r,t\rangle$, an explanation consists of the smallest set of training facts that support the prediction, as well as the rationale for each fact's inclusion in the explanation, specifically that one or more critical paths support it. 

A critical path is represented as a relation path from the head to the tail entity: $\langle h, r_1, A_1 \rangle \land \langle A_1, r_2, A_2 \rangle \land \dots \land \langle A_{n-1}, r_n, t \rangle$, where \( h \) and \( t \) represent the head and tail entities, \( r_i \) denotes relations, and \( A_i \) represents placeholder of any intermediate entity. This sequence of triples forms a path from the head to the tail entities. Each critical path corresponds to a high-confidence Closed Path (CP) rule, which describes the relationship between entities \(X\) and \(Y\) via alternative paths and consists of one or more relations without considering intermediate entities. 

\begin{figure*}[t]
\includegraphics[width=\textwidth, clip, trim=40pt 40pt 40pt 60pt]{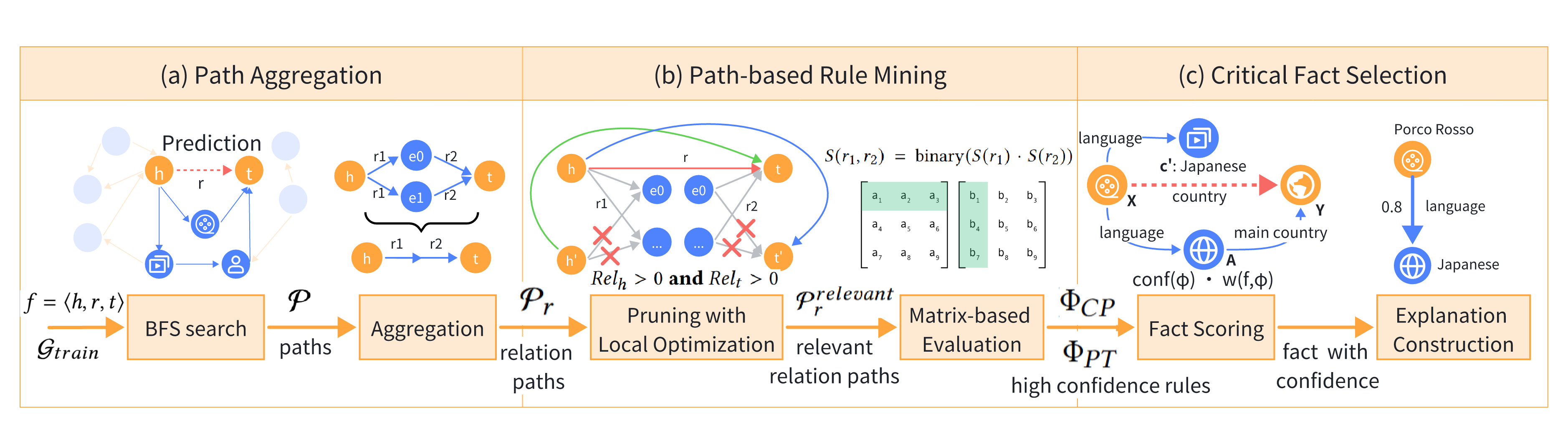}
\vspace{-0.7cm}
\caption{Pipeline of eXpath. (a) \textit{Path Aggregation}: Identifies paths between \( h \) and \( t \) using breadth-first search (BFS) and compresses them into relation paths. (b) \textit{Path-based Rule Mining}: Prunes relevant relation paths and selects high-confidence closed path (CP) and property transition (PT) rules. (c) \textit{Critical Fact Selection}: Scores candidate facts based on rule relevance and confidence, selecting the highest-scoring facts for the final explanation.}
\label{rulemining}
\vspace{-0.3cm}
\end{figure*}

Path-based explanations focus on tracing these facts and associating facts with paths. It is crucial to distinguish our path-based explanation from Power-Link~\cite{chang2024path} and PaGE-Link~\cite{zhang2023page}, which use multiple paths as explanation sets, effective in the context of GNN-based models.  These methods can leverage weighted masks to extract paths efficiently. However, in our case, embedding-based models are not inherently structured as graphs, making it difficult to directly extract paths, while the number of possible paths between a head and a tail can be enormous, and exhaustively searching this vast space is computationally impractical. Furthermore, adversarial attacks can only estimate the significance of minimal modification, while path modifications in our approach alter the dataset in a manner that is more impactful than localized changes, making it difficult to evaluate. Thus, directly using multiple paths as explanations will be less effective.


%% file: tex/4_framework.tex
\section{eXpath Method}

The eXpath method is designed to explain any given prediction $\langle h, r, t\rangle$ by identifying a small yet effective set of triples whose removal significantly impacts the model's predicted ranking of $h$ and $t$. Additionally, eXpath provides the rationale for its explanations by presenting the critical paths associated with each selected fact.

The eXpath method follows a three-stage pipeline: path aggregation, path-based rule mining, and critical fact selection. In the path aggregation stage (Figure~\ref{rulemining}(a)), we use breadth-first search (BFS) on the training facts (\( \mathcal{G}_{train} \)) to identify paths from \( h \) to \( t \), limiting the maximum path length to 3 to ensure interpretability. These paths are then compressed into relation paths ($\mathcal{P}_r$) by removing intermediate entities, reducing the candidate paths while preserving essential semantic structure. In the path-based rule mining stage (Figure~\ref{rulemining}(b)), we prune the candidate relation paths to retain only the highly relevant ones ($\mathcal{P}_r^{relevant}$) using a local optimization technique based on head and tail relevance. These relevant paths form the body of candidate closed path (CP) rules, evaluated with a matrix-based approach to compute their confidence. Simultaneously, we construct Property Transition (PT) rules from the facts linked to the head and tail entities in \( \mathcal{F}_{train}^h \) and \( \mathcal{F}_{train}^t \), retaining high-confidence CP and PT rules for fact selection. Finally, in the critical fact selection stage (Figure~\ref{rulemining}(a)), we score the candidate facts based on the number and confidence of rules they belong to, selecting the highest-scoring facts to form the final explanation.

Notably, while our method efficiently extracts path-based explanations, experiments (Section 5) show that not all KGLP explanations require path-based semantics. In sparser KGs, simple one-hop links can score higher in evaluations. To leverage both approaches, we propose a fusion model that combines eXpath's explanations with those from non-path methods (e.g., Kelpie). By evaluating explanations from both methods, the highest-scoring ones are selected as the final explanation. This fusion model highlights the complementary strengths of different explanation types and demonstrates its potential as a superior overall solution.

\subsection{Relation Path and Ontological Rules}

When providing path-based explanations for a prediction \( f = \langle h, r, t \rangle \), the number of simple paths from $h$ to $t$ grows exponentially with the path length, making even 3-hop paths computationally prohibitive. To mitigate this issue, we focus not on the specific entities traversed by a path but rather on the sequence of relations along the path. This abstraction, referred to as a "relation path," drastically reduces the number of candidate paths while preserving their semantic meaning. This concept is inspired by using \textit{closed path rules} (CP) in ontological rule learning. By aggregating multiple simple paths into relation paths, we significantly reduce path count while retaining the interpretability crucial for explanations.

\begin{figure}[t]
\includegraphics[width=\columnwidth, clip, trim=40pt 40pt 40pt 60pt]{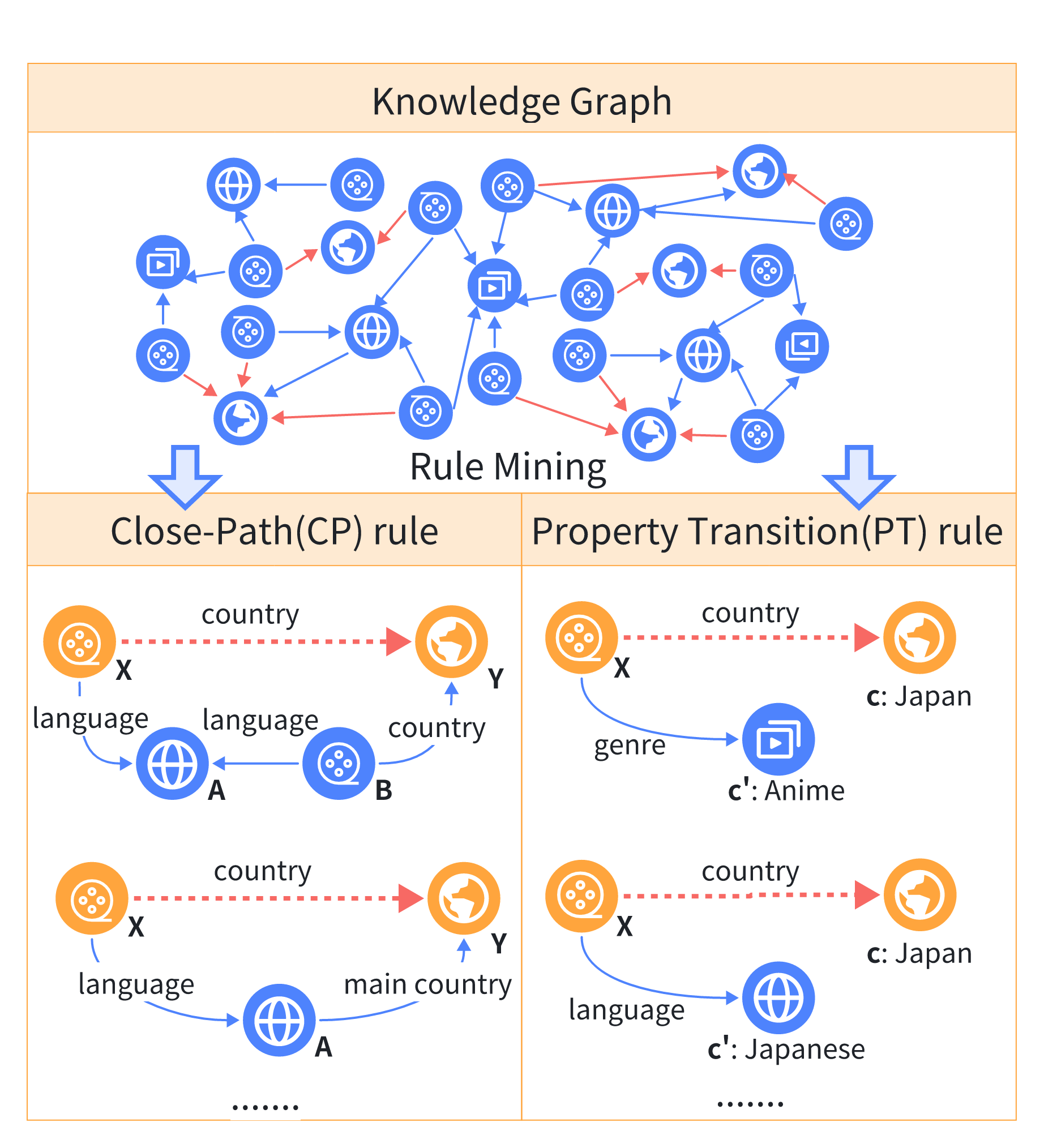}
\vspace{-0.5cm}
\caption{Principles and instances of ontological rules used in our framework. closed path (CP) rules describe the relationship between entities \(X\) and \(Y\) through alternative paths, while Property Transition (PT) rules capture transitions between different attributes of the same entity. These ontological rules are not predefined but are generalized patterns mined from the knowledge graph, supported by substructures that conform to the specified patterns.}
\label{ruleExample}
\vspace{-0.5cm}
\end{figure}

Figure~\ref{ruleExample} illustrates examples of CP and PT rules, which are inspired by the definitions of binary and unary rules with an atom ending in a constant in ontological rule mining. While PT rules can be generalized into CP rules by adding relationships between constants and replacing constants with variables, they remain essential for scenarios where two constant entities are strongly correlated (e.g., \texttt{male} and \texttt{female}) but cannot be described by simple paths. These interpretable rules offer insight into link predictions, providing a solid foundation for generating explanations. Formally, we distinguish between two types of rules:
\begin{equation}
\label{eq:rule}
\begin{aligned}
\mathbf{CP}: & \quad r\left(A_0, A_n\right) \leftarrow \bigwedge_{i=1}^n r_i\left(A_{i-1}, A_i\right) \\
\mathbf{PT}: & \quad r\left(X, c\right) \leftarrow r_0\left(X, c^{\prime}\right)\quad \text{or} \quad r\left(c, Y\right) \leftarrow r_0\left(c^{\prime}, Y\right)
\end{aligned}
\end{equation}
\noindent where \(r\) and \(r_i\) denote relations (binary predicates), \(A_0, A_i, A_n, X, Y\) are variables, and \(c, c^{\prime}\) are constants (entities). We use \(\phi\) to denote a rule, where the atoms on the left (\(h\)) form the \textit{head} of the rule (\(head(\phi)\)), and the atoms on the right (\(r\)) form the \textit{body} of the rule (\(body(\phi)\)). To simplify the notation, in the following part, we use $r\leftarrow r_1,r_2,...,r_n$ to symbolize CP rules, and relations can be reversed to capture inverse semantics (noted with a single quote, \(r' \)). For example, the relation \texttt{hypernym(X, Y)} can also be expressed as \texttt{hypernym' (Y, X)}.

CP rules are termed "closed paths" because the sequence of relations in the rule body forms a path that directly connects the subject and object arguments of the head relation. This characteristic establishes a strong connection between CP rules and relation paths. Both concepts focus on capturing the structured relationships between entities in a knowledge graph, and their forms are inherently aligned. This alignment allows relation paths to serve as direct candidates for CP rule bodies. In fact, every CP rule can be viewed as a formalized and generalized representation of a relation path, enriched with additional confidence and support. Moreover, the structured nature of CP rules makes them well-suited for explaining embedding-based predictions, as they encapsulate the critical relational patterns that underpin the model's reasoning.

To assess the quality of rules, we recall measures used in some major approaches to rule learning~\cite{chen2016scalekb, galarraga2015fast}. Let $\phi$ be a CP rule of the form ~\ref{eq:rule}. A pair of entities $r\left(e, e'\right)$ satisfies the head of $\phi$ and there exist entities $e_1, \ldots, e_{n-1}$ in the KG such that $\langle e,r_1,e_1\rangle, \ldots, \langle e_{n-1},r_n,e'\rangle$ are facts in the KG, so the body of $R$ are satisfied. Then, the support degree (supp), standard confidence (SC), and head coverage (HC) of $\phi$ are defined as:

\begin{equation}
\begin{aligned}
    \label{eq:supp}
    \operatorname{supp}(\phi)&=\#\left(e, e^{\prime}\right): \operatorname{body}(\phi)\left(e, e^{\prime}\right) \wedge r\left(e, e^{\prime}\right) \\
    S C(\phi)=&\frac{\operatorname{supp}(\phi)}{\#\left(e, e^{\prime}\right): \operatorname{body}(\phi)\left(e, e^{\prime}\right)}, H C(r)=\frac{\operatorname{supp}(\phi)}{\#(e, e): r\left(e, e^{\prime}\right)}
\end{aligned}
\end{equation}

\subsection{Path-based Rule Mining}

\begin{algorithm}[!h]
    \caption{Path-based Rule Mining Algorithm}
    \label{alg:rule_mining}
    \renewcommand{\algorithmicrequire}{\textbf{Input:}}
    \renewcommand{\algorithmicensure}{\text
    bf{Output:}}
    \begin{algorithmic}[1]
        \REQUIRE Prediction $f = \langle h,r,t\rangle$,
                 Facts from Training Set $\mathcal{G}_{train}$
        \ENSURE Candidate Rule Set for Prediction $\Phi$

        \STATE $\Phi \leftarrow \emptyset$
        \STATE \COMMENT{Step 1: CP Rule Extraction}
        \STATE $\mathcal{P} \leftarrow \text{BFSSearch}(h, t)$
        \STATE $\mathcal{P}_{r} \leftarrow \text{Aggregation}(P)$
        \FOR{each $p$ in $\mathcal{P}_{r}$}
            \STATE $h, h', t, t' \leftarrow \text{localOptimization}(f, p, \mathcal{G}_{train})$
            \STATE $Rel_h\leftarrow 1 - \frac{f_r(h', t)}{f_r(h, t)},\quad Rel_t\leftarrow 1 - \frac{f_r(h, t')}{f_r(h, t)}$
            \IF{$Rel_h > 0$ \textbf{and} $Rel_t > 0$}
                \STATE $(HC, SC, supp) \leftarrow \text{RuleEvaluation}(r\leftarrow p, \mathcal{G}_{train})$
                \IF{$SC \geq minSC$ \textbf{and} $HC \geq minHC$}
                    \STATE $\Phi \leftarrow \Phi \cup \{\phi_{CP}: r \leftarrow p [SC \times \frac{supp}{supp+minSupp}] \}$
                \ENDIF
            \ENDIF
        \ENDFOR

        \STATE \COMMENT{Step 2: PT Rule Extraction (Take Head PT Rule as Example)}
        \STATE $\mathcal{F}_{train}^h \leftarrow \text{SearchFacts}(h, \mathcal{G}_{train})$
        \FOR{each $\langle h,r_0,t_0\rangle$ in $\mathcal{F}_{train}^h$}
            \STATE $(HC, SC, supp) \leftarrow \text{RuleEvaluation}(r(X, t)\leftarrow r_0(X, t_0), \mathcal{G})$
            \IF{$SC \geq minSC$ \textbf{and} $HC \geq minHC$}
                \STATE $\Phi \leftarrow \Phi \cup \{\phi_{PT}: r(X, t) \leftarrow r_0(X, t_0) [SC \times \frac{supp}{supp+minSupp}] \}$
            \ENDIF
        \ENDFOR

        \RETURN $\Phi$
    \end{algorithmic}
\end{algorithm}

A critical step for generating path-based explanations is constructing a rule set $\Phi$, which includes both closed path (CP) and Property Transition (PT) rules, as defined in Section 4.1. We do not mine all possible rules across the entire knowledge graph (KG) but instead focus on extracting relevant rules for each prediction from a localized graph relevant to the specific prediction $f=\langle h, r, t \rangle$.

PT rules relevant to a given prediction arise from other facts related to $h$ and $t$ ($f' \in \mathcal{F}_{train}^h \cup \mathcal{F}_{train}^t$). These rules are constructed by replacing common entities in $f$ and $f'$ with variables, which serve as the rule head and body, respectively. For example, for $f = \langle$Porco\_Rosso, language, Japanese$\rangle$ and $f' = \langle$Porco\_Rosso, genre, Anime$\rangle$, the corresponding PT rule is: $\langle$X, language, Japanese$\rangle \leftarrow \langle X, \text{genre}, \text{Anime} \rangle$. This rule, similar to the "sufficient scenario" proposed by Kelpie~\cite{rossi2022explaining}, captures whether different entities in the same context satisfy the same prediction.

Calculating metrics for PT rules is relatively straightforward. Based on Equation~\ref{eq:supp}, we simply count the number of facts in $\mathcal{G}_{train}$ that satisfy $\langle X, \text{language}, \text{Japanese} \rangle$ and $\langle X, \text{genre}, \text{Anime} \rangle$ as the head and body counts, respectively. The number of facts satisfying both conditions serves as the support count. Finally, we set a threshold: only rules for which $SC(\phi) > \text{minSC}$ and $HC(\phi) > \text{minHC}$ are selected to form the PT rule set $\Phi_{PT}$.

CP rules relevant to a prediction, on the other hand, arise from relation paths ($\mathcal{P}_r$) connecting $h$ and $t$. CP rule mining is more complex than PT rule mining due to the potentially large number of CP rules for a single prediction and the computational expense of evaluating CP rules across the entire knowledge graph. As detailed in Algorithm~\ref{alg:rule_mining}, we first filter $\mathcal{P}_r$ using local optimization, ensuring that only relation paths relevant to the prediction $\mathcal{P}_r^{relevant}$ are considered for evaluation.

During the pruning process, each relation path is assigned a head relevance score and a tail relevance score, which reflect its importance to the prediction. Relation paths with positive head and tail relevance ($Rel_h>0$ and $Rel_t>0$) scores are considered relevant to the prediction and retained as candidate rule bodies ($\mathcal{P}_r^{relevant}$) for further evaluation. This filtering approach assumes that a relation path can only serve as a valid rule body if both its head and tail relations are critical to the prediction.

To compute relevance scores, eXpath adopts an efficient local optimization approach inspired by the Kelpie mimic strategy~\cite{rossi2022explaining}. Mimic entities for the head and tail, denoted as $h'$ and $t'$ (see Fig.~\ref{rulemining}(b)), are created. These mimic entities retain the same connections as the original head or tail entities, except that all facts associated with the evaluated relation are removed. The embeddings of the mimic entities, along with those of the original head and tail entities, are then independently trained using their directly connected facts.

Three predictive scores are computed: $f_r(h, t)$, $f_r(h', t)$, and $f_r(h, t')$, where $f_r(h, t)$ represents the model’s scoring function for the triple $\langle h, r, t \rangle$. The relevance of a relation is defined as the reduction in the predictive score after removing all facts associated with a specific relation:

\begin{equation}
    \label{eq:relevance}
    Rel_h = 1 - \frac{f_r(h', t)}{f_r(h, t)}, \quad Rel_t = 1 - \frac{f_r(h, t')}{f_r(h, t)}
\end{equation}

Here, $Rel_h$ and $Rel_t$ quantify the importance of relations connected to the head and tail entities. Relative changes in scores are used instead of rank reductions, as scores provide a more robust metric. Rank reductions can be unreliable, especially in local optimization scenarios where mimic entities may overfit, resulting in consistent ranks of 1. This relevance score effectively captures the impact of facts on the prediction by simulating the model's underlying embedding mechanisms.

Finally, eXpath constructs a CP rule set $\Phi_{CP}$ for each prediction based on the relevant relation paths $\mathcal{P}_r^{relevant}$ to select high-quality rules that have strong support and confidence. Confidence is computed as $conf(\phi) = SC(\phi) \cdot \frac{supp(\phi)}{supp(\phi) + \text{minSupp}}$, which prevents the overestimation of rules with insufficient support (e.g., $supp < 10$), inadequate for generalizing into a rule. High-confidence CP and PT rules ($\Phi_{CP}$ and $\Phi_{PT}$) are retained for fact selection. Strong support and confidence ensure that the selected rules are robust for causal reasoning, enabling eXpath to generate accurate and interpretable path-based explanations.

However, efficiently computing metrics for CP rules presents a significant challenge. To address this, we adopt the matrix-based approach from RLvLR~\cite{omran2019embedding}. The method verifies the satisfiability of the body atoms in candidate rules to compute the metrics for CP rules. Given a KG represented as a set of \( S \) matrices, where each \( n \times n \) binary matrix $S(r_k)$ corresponds to a relation $r_k$, the adjacency matrix $S(r_k)$ has an entry of 1 if the fact $\langle e_i, r_k, e_j \rangle$ exists in the KG, and 0 otherwise.

The product of adjacency matrices is closely related to closed path rules. For instance, consider the rule $\phi: r \leftarrow r_1, r_2$. A fact $r_t(e, e')$ is inferred by $\phi$ if there exists an entity $e''$ such that $r_1(e, e'')$ and $r_2(e'', e')$ hold. The product $S(r_1) \cdot S(r_2)$ produces the adjacency matrix for the set of inferred facts. The binary transformation $S(r_1, r_2) = \text{binary}(S(r_1) \cdot S(r_2))$ is then used to generalize this computation. The metrics for this CP rule are calculated as:

\begin{equation}
    \begin{aligned}
    \text{supp}(\phi) = \text{sum}(S(r_1, r_2) \& S(r)) \\
    \text{SC}(\phi) = \frac{\text{supp}(\phi)}{\text{sum}(S(r_1, r_2))}, \text{HC}(\phi) &= \frac{\text{supp}(\phi)}{\text{sum}(S(r))}
    \end{aligned}
\end{equation}

\noindent where $\text{sum}$ aggregates all matrix entries, and $\&$ represents the element-wise logical AND operation. While this example involves rules with its body of length 2, the method extends straightforwardly to any length. This matrix-based approach offers a scalable solution for efficiently computing rule metrics in large knowledge graphs.

\subsection{Critical Fact Selection}

This section details the method for selecting an optimal set of facts to explain a given prediction triple \( \langle h, r, t \rangle \), leveraging the rules extracted in the previous step. The core idea is to identify the most critical fact or a combination of facts within the paths connecting the head and tail entities. Each fact is evaluated based on its contribution to the prediction, and those with higher scores are considered more pivotal. The final explanation set is constructed by selecting the highest-scoring facts.

Several key factors are taken into account to determine the significance of a fact: (1) Facts that satisfy a larger number of rules are given higher priority, as this indicates their broader relevance within the prediction. (2) Rules with higher confidence are weighted more heavily, reflecting their more robust causal support. (3) The position of a fact within a rule (e.g., whether it connects to the head or tail entity) is adjusted based on the relation relevance scores determined earlier.

Considering all these factors, the scoring system provides a robust metric for evaluating each fact's importance. To model the contribution of a fact that satisfies multiple rules, we adopt a confidence degree (CD) aggregation approach inspired by rule-based link prediction methods~\cite{omran2018scalable}. The CD of a fact \( f \) is calculated using the confidence values of all the rules that infer \( f \) in a Noisy-OR manner. For explanation tasks, which reverse the link prediction perspective, we define the CD of \( f \) as follows:

\begin{equation}
    C D(f) = 1 - \prod_{\phi \in \Phi(f)}(1 - conf(\phi) \cdot w(f, \phi))
\end{equation}

\noindent where \( \Phi(f) \) is the set of rules inferred from the prediction, \( conf(\phi) \) is the confidence of rule \( \phi \), and \( w(f, \phi) \) represents the importance of fact \( f \) within rule \( \phi \). This importance score, ranging from 0 to 1, reflects the proportion of \( f \) 's appearances in the rule and its relative importance based on the relevance of the rule's head and tail relations. The importance score \( w(f, \phi) \) is calculated as:

\begin{equation}
    \begin{aligned}
r_h(\phi) &= \frac{Rel_h(\phi)}{Rel_h(\phi) + Rel_t(\phi)}\\
w(f, \phi) &= r_h(\phi) \cdot p_h(f, \phi) + (1 - r_h(\phi)) \cdot p_t(f, \phi)
\end{aligned}
\end{equation}

\noindent where \( Rel_h(\phi) \) and \( Rel_t(\phi) \) are the relevance scores of the rule's head and tail relations, respectively. The term \( p_h(f, \phi) \) represents the proportion of \( f \) 's appearances in the head of all paths related to rule $\phi$. This formulation ensures that facts appearing more prominently in rules are scored higher. In PT rules, the importance score for a fact \(w(f, \phi)\) is simplified to 1, as the rule corresponds to a unique fact for a given prediction.

We rank all candidate facts by their CD scores and select the top-ranked facts to form the explanation. This approach ensures that the selected facts are those most strongly supported by high-quality, relevant rules, providing robust and interpretable explanations for the given prediction.

%% file: tex/5_experiment.tex
\section{Experiment}
\subsection{Experimental Setup}

We assessed eXpath on the KG LP task using four benchmark datasets: FB15k, FB15k-237 ~\cite{lacroix2018canonical}, WN18, and WN18RR ~\cite{bhowmik2020explainable}. These datasets' detailed statistics and link prediction metrics are provided in Table~\ref{tab:dataset}. We adhered to the standard splits and training parameters to ensure consistency across comparisons and maintain identical training parameters before and after removing facts.


\begin{table}[t]
\centering
\caption{Statistics of benchmark datasets.}
\vspace{-0.3cm}
\setlength{\tabcolsep}{5pt}
\renewcommand{\arraystretch}{1.2}
\begin{tabular}{lrrrrr}
\toprule
\makecell[c]{KG\\Dataset} & 
\makecell[c]{Entities} & 
\makecell[c]{Relation\\Types} & 
\makecell[c]{Train\\Facts} & 
\makecell[c]{Valid\\Facts} & 
\makecell[c]{Test\\Facts} \\
\midrule
FB15k       & 14,951   & 1,345          & 483,142     & 50,000      & 50,971     \\
FB15k-237   & 14,541   & 237            & 272,115     & 17,535      & 20,466     \\
WN18        & 40,943   & 18             & 141,442     & 5,000       & 5,000      \\
WN18RR      & 40,943   & 11             & 86,835      & 3,034       & 3,134      \\
\bottomrule
\vspace{-0.5cm}
\end{tabular}
\label{tab:dataset}
\end{table}

We compared the performance of eXpath against four contemporary systems dedicated to LP interpretation: Kelpie~\cite{rossi2022explaining}, Data Poisoning (DP)~\cite{zhang2019data}, Criage~\cite{betz2022adversarial}, and KGEAttack~\cite{betz2022adversarial}. These implementations are publicly available, and we tailored the code sourced from their respective Github repositories. Since the explanation framework is compatible with any Link Prediction (LP) model rooted in embeddings, we conduct experiments on three models with different loss functions: CompEx~\cite{trouillon2016complex}, ConvE~\cite{dettmers2018convolutional}, and TransE~\cite{wang2014knowledge}.

In adversarial attacks, each explanation framework recommends one or more facts, which are removed before retraining the model with the same parameters. The drop in performance metrics is used to assess the quality of the explanations. The baseline frameworks, including DP, Criage, and Kelpie, focus solely on facts directly related to the head entity (i.e., attributes of the head entity). KGEAttack randomly selects a fact in the extracted rule, while eXpath focuses on facts related to either the head or tail entity, each supported by relevant CP and PT rules. To ensure fairness between the explanation systems, we restrict the number of facts that can be removed. Specifically, DP, Criage, Kelpie(L1), and eXpath(L1) limit the removal to at most one fact, whereas Kelpie and eXpath can remove up to four facts. Based on experiments and existing literature, we set the thresholds \( \text{minSC} = 0.1, \text{minHC} = 0.01, \text{minSupp} = 10 \). These parameters are adapted from the definitions of high-quality rules in prior work~\cite{galarraga2015fast}.

Based on the problem formulation outlined in Section 3.3, we randomly select a small subset \( T \subset \mathcal{G}_e \) from the test set, where the model demonstrates relatively good predictive performance. Specifically, we choose 100 predictions that exhibit strong performance. These predictions are not required to rank first for both head and tail predictions, as enforcing such strict criteria could overly limit the selection process and reduce the applicability of the scenarios. To evaluate model performance, we focus on the relative reduction in reciprocal rank rather than the absolute reduction since the predictions in \( T \) are not necessarily top-ranked, and lower-performing predictions are assigned smaller weights. The model's explanatory capability is measured by the relative reduction in H@1 (Hits@1) and MRR (Mean Reciprocal Rank), defined as:

\begin{equation}
\label{eq:delta_mrr}
\begin{aligned}
H@1(M_x,f) &=\frac{1}{2}(\mathbf{1}(rk_h(M_x,f)=1) + \mathbf{1}(rk_t(M_x,f)=1)) \\
RR(M_x, f) &= \frac{1}{2} \left( \frac{1}{rk_h(M_x, f)} + \frac{1}{rk_t(M_x, f)} \right) \\
\delta H@1(M_x, T) &= 1 - \frac{\sum_{f\in T} H@1(M_x,f)}{\sum_{f\in T} H@1(M_o,f)} \\
\delta MRR(M_x, T) &= 1 - \frac{\sum_{f\in T} RR(M_x,f)}{\sum_{f\in T} RR(M_o,f)}
\end{aligned}
\end{equation}

\noindent where \( M_x \) represents the model trained on the dataset excluding the candidate explanations extracted by the explanation framework \( x \), and \( M_o \) denotes the original model trained on the entire dataset, $\mathbf{1}(\cdot)$  is the indicator function that returns 1 if the condition inside holds and 0 otherwise. 

While both $\delta H@1$ and $\delta MRR$ are useful, $\delta MRR$ proves more robust. The stochasticity of model training and small dataset size (100 predictions) can cause significant variability in $\delta H@1$ values. This issue is exacerbated for fragile models like TransE, where ranks fluctuate even without attacks. We address this by averaging results over five experimental runs. To ensure that the prediction to be explained is of high quality, we restrict the MRR to greater than 0.5, which ensures that the prediction ranks first in at least one of the head or tail predictions. To evaluate the overall explanatory power, we sum the MRR values of new model \( M_x \) and original model \( M_o \) across all facts in the numerator and denominator, respectively. This approach ensures that better predictions contribute more significantly to the evaluation, avoiding bias toward selecting only predictions with head and tail ranks 1. Moreover, this method allows for negative explanations, where the rank decreases after removing a fact.

\subsection{Explanation Results}

\begin{table*}[h!]
\centering
\caption{$\delta H@1$ comparison across different models and datasets using various explanation methods. All results are averaged over five runs, with higher values indicating better performance. The original $H@1$ is 1 for all candidate predictions ($H@1 > 1$ predictions are excluded). Methods with "+eXpath" indicate fusion approaches that combine the given method with eXpath.}
\vspace{-0.2cm}
\label{tab:deltaH1}
\begin{tabular}{lcccccccccccccc}
\toprule
\multirow{2}{*}{\makecell[c]{Max \\ Exp. \\ Size}} & \multirow{2}{*}{Method} & \multicolumn{4}{c}{Complex} & \multicolumn{4}{c}{Conve} & \multicolumn{4}{c}{TransE} & \multirow{2}{*}{AVG} \\
\cmidrule(lr){3-6} \cmidrule(lr){7-10} \cmidrule(lr){11-14}
 & & \rotatebox{60}{FB15k} & \rotatebox{60}{FB15k-237} & \rotatebox{60}{WN18} & \rotatebox{60}{WN18RR} 
 & \rotatebox{60}{FB15k} & \rotatebox{60}{FB15k-237} & \rotatebox{60}{WN18} & \rotatebox{60}{WN18RR} 
 & \rotatebox{60}{FB15k} & \rotatebox{60}{FB15k-237} & \rotatebox{60}{WN18} & \rotatebox{60}{WN18RR} & \\
\midrule
\multirow{9}{*}{\makecell[c]{single\\-fact\\exp.}} & Criage~\cite{pezeshkpour2019investigating} & .087 & .105 & .080 & .203 & .153 & .162 & .270 & .256 & --- & --- & --- & --- & .165 \\
 & DP~\cite{zhang2019data} & .529 & .315 & .799 & .758 & .246 & .162 & .794 & .829 & .304 & .326 & .910 & .709 & .557 \\
 & Kelpie~\cite{rossi2022explaining} & .576 & .395 & .578 & .593 & .229 & .222 & .567 & .667 & .261 & .281 & .792 & .779 & .495 \\
 & KGEAttack~\cite{betz2022adversarial} & .547 & .290 & .829 & .764 & .237 & .212 & .929 & .915 & .365 & .213 & .938 & .779 & .585 \\
 & eXpath & .512 & .395 & .834 & .797 & .271 & .343 & .929 & .891 & .313 & .337 & .938 & .767 & .611 \\
 & Criage+eXPath & .523 & .411 & .839 & .819 & .322 & .404 & \textbf{.936} & .891 & --- & --- & --- & --- & .643 (+290\%)  \\
 & DP+eXpath & .570 & .500 & \textbf{.859} & .813 & .331 & .414 & \textbf{.936} & \textbf{.946} & .374 & \textbf{.438} & \textbf{.944} & .826 & .663 (+19\%) \\
 & Kelpie+eXPath & \textbf{.657} & \textbf{.540} & \textbf{.859} & \textbf{.835} & \textbf{.364} & \textbf{.424} & .929 & .915 & \textbf{.417} & .427 & \textbf{.944} & \textbf{.872} & \textbf{.682} (+38\%) \\
 & KGEA.+eXpath & .576 & .452 & \textbf{.859} & .802 & .322 & .384 & .929 & \textbf{.946} & \textbf{.417} & .360 & .938 & \textbf{.872} & .655 (+12\%) \\
\midrule
\multirow{3}{*}{\makecell[c]{four\\-fact\\exp.}} & Kelpie & .767 & .581 & .829 & .940 & .534 & .303 & .816 & .946 & .374 & .427 & .868 & .907 & .691 \\
 & eXpath & .802 & .661 & .920 & .951 & .542 & .566 & .957 & .984 & .539 & .573 & \textbf{.965} & \textbf{.965} & .785 \\
 & Kelpie+eXpath & \textbf{.831} & \textbf{.742} & \textbf{.935} & \textbf{.989} & \textbf{.653} & \textbf{.596} & \textbf{.965} & \textbf{.984} & \textbf{.609} & \textbf{.674} & \textbf{.965} & \textbf{.965} & \textbf{.826} \\
\bottomrule
\end{tabular}
\end{table*}

\begin{table*}[h!]
\centering
\caption{$\delta MRR$ comparison across different models and datasets using various explanation methods. All results are averaged over five runs, with higher values indicating better performance. The original MRR is above 0.5 in all candidate predictions.}
\vspace{-0.2cm}
\label{tab:deltaMRR}
\begin{tabular}{lcccccccccccccc}
\toprule
\multirow{2}{*}{\makecell[c]{Max \\ Exp. \\ Size}} & \multirow{2}{*}{Method} & \multicolumn{4}{c}{Complex} & \multicolumn{4}{c}{Conve} & \multicolumn{4}{c}{TransE} & \multirow{2}{*}{AVG} \\
\cmidrule(lr){3-6} \cmidrule(lr){7-10} \cmidrule(lr){11-14}
 & & \rotatebox{60}{FB15k} & \rotatebox{60}{FB15k-237} & \rotatebox{60}{WN18} & \rotatebox{60}{WN18RR} 
 & \rotatebox{60}{FB15k} & \rotatebox{60}{FB15k-237} & \rotatebox{60}{WN18} & \rotatebox{60}{WN18RR} 
 & \rotatebox{60}{FB15k} & \rotatebox{60}{FB15k-237} & \rotatebox{60}{WN18} & \rotatebox{60}{WN18RR} & \\
\midrule
\multirow{9}{*}{\makecell[c]{single\\-fact\\exp.}} & Criage & .045 & .051 & .058 & .163 & .024 & .031 & .157 & .150 & --- & --- & --- & --- & .085 \\
 & DP & .451 & .187 & .729 & .668 & .140 & .058 & .728 & .785 & .157 & .141 & .742 & .613 & .450 \\
 & Kelpie & .457 & .238 & .491 & .483 & .123 & .076 & .514 & .578 & .075 & .115 & .700 & .664 & .376 \\
 & KGEAttack & .463 & .172 & .766 & .684 & .159 & .104 & .889 & .853 & .190 & .091 & .877 & .659 & .492 \\
 & eXpath & .430 & .233 & .774 & .688 & .183 & .130 & .889 & .810 & .159 & .165 & .877 & .596 & .494 \\
 & Criage+eXpath & .443 & .236 & .777 & .711 & .203 & .185 & .892 & .814 & --- & --- & --- & --- & .533 (+527\%) \\
 & DP+eXpath & .491 & .282 & \textbf{.803} & .711 & .241 & .211 & \textbf{.900} & \textbf{.893} & .239 & \textbf{.252} & .891 & .675 & .549 (+22\%) \\
 & Kelpie+eXpath & \textbf{.534} & \textbf{.309} & .795 & \textbf{.718} & \textbf{.245} & .206 & .895 & .848 & .225 & .239 & \textbf{.893} & \textbf{.734} & \textbf{.553} (+47\%) \\
 & KGEA.+eXpath & .495 & .262 & .799 & .712 & .239 & \textbf{.215} & .889 & .883 & \textbf{.261} & .223 & .877 & .723 & .548 (+12\%) \\
\midrule
\multirow{3}{*}{\makecell[c]{four\\-fact\\exp.}} & Kelpie & .632 & .434 & .777 & .891 & .391 & .143 & .795 & .919 & .203 & .199 & .805 & .893 & .590 \\
 & eXpath & .680 & .452 & .875 & .887 & .366 & .327 & .924 & .952 & .354 & .261 & .937 & .943 & .663 \\
 & Kelpie+eXpath & \textbf{.718} & \textbf{.519} & \textbf{.900} & \textbf{.941} & \textbf{.468} & \textbf{.401} & \textbf{.949} & \textbf{.966} & \textbf{.406} & \textbf{.332} & \textbf{.952} & \textbf{.960} & \textbf{.709} \\
\bottomrule
\end{tabular}
\end{table*}

Tables ~\ref{tab:deltaH1} and ~\ref{tab:deltaMRR} demonstrate the overall effectiveness of the eXpath method in generating explanations for link prediction tasks, evaluated using the $\delta H@1$ and $\delta MRR$ metrics as defined in Equation~\ref{eq:delta_mrr}. For a fair comparison, explanation methods are categorized based on explanation size (i.e., the number of facts provided). The first section of each table (top 9 rows) presents results for five single-fact explanations (L1) and their fusion models, such as Criage, DP, Kelpie, KGEAttack, and eXpath, which offer one fact per explanation. The second section (bottom 3 rows) shows results for four-fact explanations (L4), including eXpath, Kelpie, and their fusion.

For single-fact explanations, eXpath achieves the best average performance, with an average of 0.611 in \( \delta H@1 \) and 0.494 in \( \delta MRR \). KGEAttack performs comparably, reaching an average of 0.585 in \( \delta H@1 \) and 0.492 in \( \delta MRR \). Both methods significantly outperform Criage and Kelpie, surpassing them by at least 15.4\% in \( \delta H@1 \) and 23.6\% in \( \delta MRR \) on average. Notably, eXpath secures at least the second-best performance in 20 out of 24 settings and significantly outperforms all methods in 12 settings. Interestingly, eXpath explanations exhibit dataset-specific preferences. Compared to KGEAttack, eXpath performs better in explaining relation-dense datasets such as FB15k-237, achieving an average improvement of 50.3\% in \( \delta H@1 \) and 43.8\% in \( \delta MRR \). On other datasets, the performance of both methods is similar.

In a more practical four-fact scenario, only eXpath and Kelpie support multiple facts as explanations. eXpath, which directly selects the top-scoring set of up to four facts, outperforms Kelpie in 22 out of 24 settings with statistical significance ($p$-value < 0.05) across five runs. Specifically, eXpath achieves an average of 0.785 in $\delta H@1$ and 0.663 in $\delta MRR$, while Kelpie achieves averages of 0.691 in $\delta H@1$ and 0.590 in $\delta MRR$. Notably, four-fact explanations of eXpath consistently outperform single-fact explanations across all settings, emphasizing the importance of multi-fact combinations for meaningful explanations. This is particularly evident in dense datasets like FB15k and FB15k-237, where four-fact explanations show an average improvement of 69.5\% in $ \delta H@1$ and 87.7\% in $ \delta MRR$, compared to single-fact explanations. In contrast, for sparser datasets like WN18 and WN18RR, the improvements are more modest, with average gains of 11.3\% in $ \delta H@1$ and 41.4\% in $ \delta MRR$. Dense graphs, such as FB15k, contain many synonyms or antonyms for relations (e.g., \texttt{actor-film}, \texttt{sequel-prequel}, \texttt{award-honor}), meaning that even if one fact is removed from an explanation, other related facts remain in the knowledge graph, making adversarial attacks less effective. This observation underscores the need for multi-fact explanations to fully capture the predictive context.

We also evaluate fusion methods (e.g., Kelpie+eXpath), selecting the explanation that yields the greater reduction in MRR, defined as $RR(M_{x+y}, f) = \min(RR(M_x, f), RR(M_y, f))$. Fusion methods significantly enhance explanation performance. For instance, combining eXpath(L1) with Criage, DP, Kelpie(L1), and KGEAttack improves $\delta MRR$ by 527\%, 22\%, 47\%, and 12\%, respectively. The eXpath-Kelpie fusion improves Kelpie alone by 20\%. These results demonstrate that eXpath offers diverse and complementary perspectives, particularly when integrated with Kelpie, highlighting differences in explanation strategies. However, L1 fusion methods converge to an upper bound ($\delta MRR\leq 0.56, \delta H@1 \leq 0.69$), indicating that single-fact explanations have inherent limitations. Multi-fact approaches are necessary for satisfactory explanations in link prediction tasks.

\begin{figure}[t]
\centering
\includegraphics[width=\columnwidth, clip, trim=40pt 20pt 20pt 20pt]{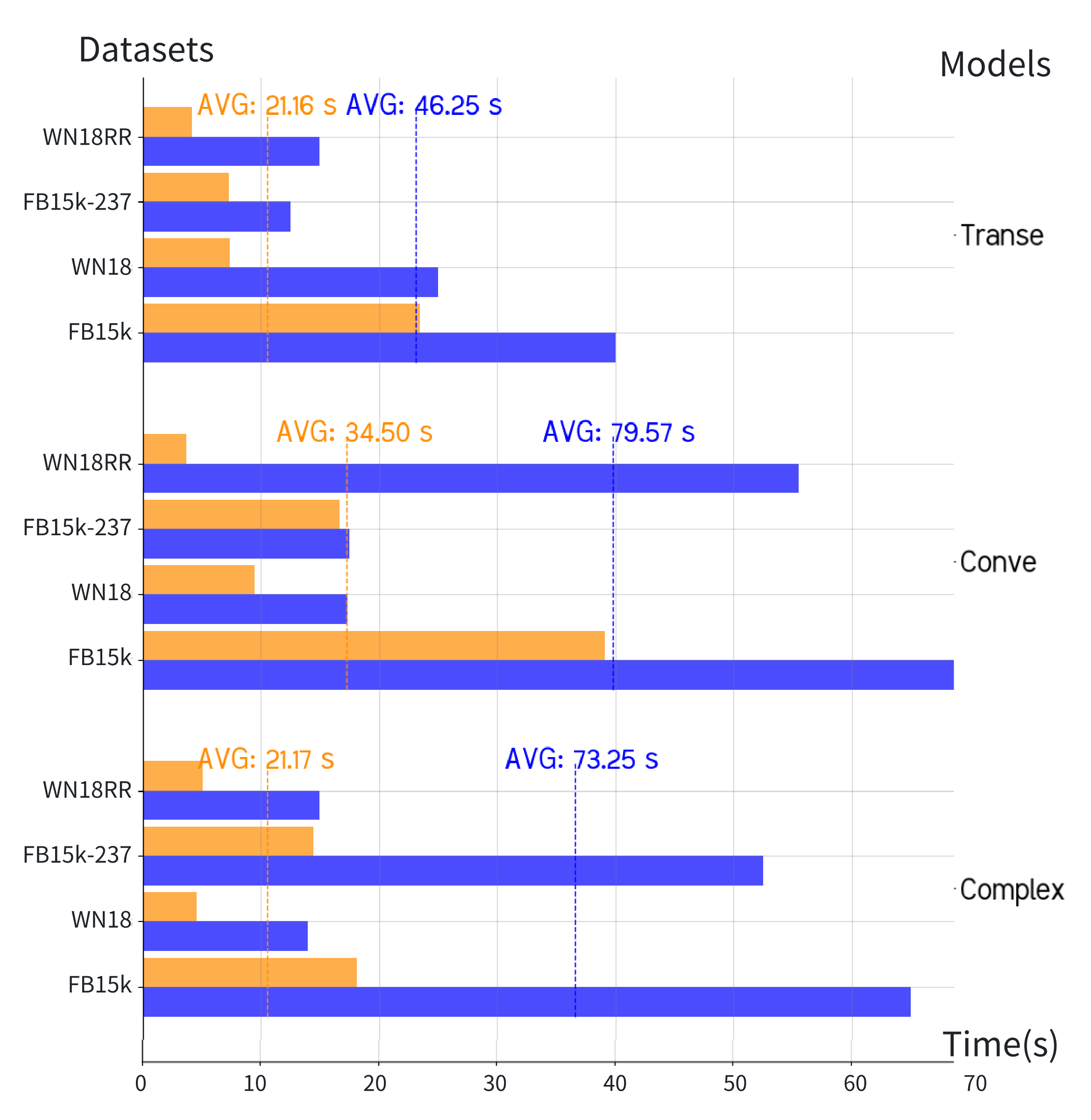}
\vspace{-0.3cm}
\caption{Average times in seconds to extract an explanation for Kelpie and eXpath.} \label{time}
\end{figure}

In terms of efficiency, Figure~\ref{time} compares the average explanation time per prediction between eXpath and Kelpie. eXpath achieves significantly faster explanation speeds, averaging 25.61 seconds per prediction, which is approximately 38.6\% of Kelpie's average time of 66.36 seconds. This efficiency is attributed to eXpath's localized optimization within relation groups and its straightforward scoring-based fact selection process, compared to Kelpie's exhaustive traversal of connections and time-intensive combinatorial searches.

In conclusion, eXpath demonstrates clear advantages in both performance and execution efficiency, highlighting its potential as a robust framework for advancing path-based and rule-based explanation systems in link prediction tasks.

\subsection{Fact Position Preferences}

\begin{table*}[h!]
\centering
\caption{$\delta MRR$ Comparison between different models and datasets with different fact position preferences: \texttt{all} denotes unrestricted fact selection, \texttt{head} restricts facts to those connected to the head entity, and \texttt{tail} restricts facts to those connected to the tail entity.}
\vspace{-0.3cm}
\label{tab:preference}
\begin{tabular}{lcccccccccccccc}
\toprule
\multirow{2}{*}{\makecell[c]{Max \\ Exp. \\ Size}} & \multirow{2}{*}{Method} & \multicolumn{4}{c}{Complex} & \multicolumn{4}{c}{Conve} & \multicolumn{4}{c}{TransE} & \multirow{2}{*}{AVG} \\
\cmidrule(lr){3-6} \cmidrule(lr){7-10} \cmidrule(lr){11-14}
 & & \rotatebox{60}{FB15k} & \rotatebox{60}{FB15k-237} & \rotatebox{60}{WN18} & \rotatebox{60}{WN18RR} 
 & \rotatebox{60}{FB15k} & \rotatebox{60}{FB15k-237} & \rotatebox{60}{WN18} & \rotatebox{60}{WN18RR} 
 & \rotatebox{60}{FB15k} & \rotatebox{60}{FB15k-237} & \rotatebox{60}{WN18} & \rotatebox{60}{WN18RR} & \\
\midrule
\multirow{3}{*}{1} 
 & eXpath(all) & .431 & .233 & \textbf{.774} & .696 & .163 & \textbf{.135} & \textbf{.889} & \textbf{.833} & \textbf{.159} & \textbf{.149} & \textbf{.877} & .406 & .479 \\
 & eXpath(head) & .433 & \textbf{.243} & \textbf{.774} & .693 & .165 & .119 & \textbf{.889} & .810 & .148 & .127 & \textbf{.877} & \textbf{.598} & \textbf{.490} \\
 & eXpath(tail) & \textbf{.448} & .125 & .759 & \textbf{.635} & \textbf{.177} & .088 & \textbf{.889} & .787 & .159 & .071 & \textbf{.877} & -.059 & .413 \\
\midrule
\multirow{3}{*}{2} 
 & eXpath(all) & .520 & \textbf{.343} & .783 & \textbf{.809} & .230 & .196 & \textbf{.888} & .900 & .231 & .177 & .904 & .602 & .549 \\
 & eXpath(head) & \textbf{.539} & .329 & .780 & .769 & .223 & .196 & \textbf{.889} & \textbf{.902} & \textbf{.254} & \textbf{.251} & \textbf{.908} & \textbf{.820} & \textbf{.572} \\
 & eXpath(tail) & .549 & .156 & \textbf{.784} & .737 & \textbf{.291} & \textbf{.134} & .885 & .869 & .227 & .134 & .857 & .000 & .469 \\
\midrule
\multirow{3}{*}{4} 
 & eXpath(all) & \textbf{.680} & \textbf{.453} & .807 & .878 & .370 & \textbf{.319} & .900 & .939 & \textbf{.355} & .270 & .918 & .826 & .643 \\
 & eXpath(head) & .659 & .438 & \textbf{.877} & \textbf{.887} & \textbf{.372} & .290 & \textbf{.925} & \textbf{.952} & .346 & \textbf{.271} & \textbf{.935} & \textbf{.942} & \textbf{.658} \\
 & eXpath(tail) & .630 & .227 & .833 & .818 & .324 & .103 & .877 & .859 & .232 & .135 & .843 & .125 & .501 \\
\midrule
\multirow{3}{*}{8} 
 & eXpath(all) & \textbf{.762} & \textbf{.584} & .850 & .956 & .449 & .419 & .930 & .961 & .471 & .333 & .932 & .909 & .713 \\
 & eXpath(head) & .727 & .558 & .904 & \textbf{.990} & .438 & .394 & .919 & \textbf{.978} & \textbf{.533} & \textbf{.350} & \textbf{.965} & \textbf{.959} & \textbf{.726} \\
 & eXpath(tail) & .737 & .337 & \textbf{.927} & .920 & \textbf{.450} & \textbf{.240} & \textbf{.927} & .913 & .389 & .201 & .923 & .367 & .611 \\
\bottomrule
\end{tabular}
\end{table*}

\begin{table*}[h!]
\centering
\caption{Ablation study results on $\delta \text{MRR}$, comparing the impact of excluding CP rules (\textbf{w/o CP}) and PT rules (\textbf{w/o PT}) across different models and datasets. The \textbf{Sparse} strategy selects facts associated with the sparsest relations as explanations, serving as a baseline for comparison. }
\vspace{-0.3cm}
\label{tab:ablation}
\begin{tabular}{lcccccccccccccc}
\toprule
\multirow{2}{*}{\makecell[c]{Max \\ Exp. \\ Size}} & \multirow{2}{*}{Method} & \multicolumn{4}{c}{Complex} & \multicolumn{4}{c}{Conve} & \multicolumn{4}{c}{TransE} & \multirow{2}{*}{AVG} \\
\cmidrule(lr){3-6} \cmidrule(lr){7-10} \cmidrule(lr){11-14}
 & & \rotatebox{60}{FB15k} & \rotatebox{60}{FB15k-237} & \rotatebox{60}{WN18} & \rotatebox{60}{WN18RR} 
 & \rotatebox{60}{FB15k} & \rotatebox{60}{FB15k-237} & \rotatebox{60}{WN18} & \rotatebox{60}{WN18RR} 
 & \rotatebox{60}{FB15k} & \rotatebox{60}{FB15k-237} & \rotatebox{60}{WN18} & \rotatebox{60}{WN18RR} & \\
\midrule
\multirow{4}{*}{1} & eXpath & .431 & .223 & .774 & .693 & .163 & .135 & .889 & .810 & .159 & .149 & .877 & .598 & .492 \\
 & eXpath (w/o CP) & \textbf{.276} & .195 & \textbf{.757} & \textbf{.659} & \textbf{.083} & \textbf{.125} & \textbf{.448} & \textbf{.423} & \textbf{.106} & \textbf{.153} & \textbf{.520} & \textbf{.574} & \textbf{.360} (-27\%) \\
 & eXpath (w/o PT) & .431 & \textbf{.118} & .774 & .685 & .154 & .047 & .889 & .853 & .155 & .097 & .877 & .558 & .470 (-4.5\%) \\
\midrule
\multirow{4}{*}{4} & eXpath & .680 & .453 & .877 & .887 & .370 & .319 & .925 & .952 & .355 & .270 & .935 & .942 & .664 \\
 & eXpath (w/o CP) & \textbf{.477} & .416 & \textbf{.875} & .877 & \textbf{.212} & .295 & \textbf{.708} & \textbf{.835} & \textbf{.190} & \textbf{.276} & \textbf{.800} & \textbf{.936} & \textbf{.575} (-13.5\%) \\
 & eXpath (w/o PT) & .622 & \textbf{.305} & .833 & \textbf{.839} & .341 & \textbf{.159} & .925 & .953 & .329 & .174 & .941 & .930 & .613 (-6\%) \\
\bottomrule
\end{tabular}
\end{table*}



To evaluate the effect of restricting facts to the head or tail entity, we analyzed their impact on explanation performance, focusing on the relative significance of head and tail attributes. Table~\ref{tab:preference} presents results across explanation sizes (1, 2, 4, 8), where \texttt{all} allows unrestricted fact selection, \texttt{head} restricts facts to those connected to the head entity, and \texttt{tail} restricts facts to those connected to the tail entity. Facts unrelated to the head or tail are excluded as they do not directly influence embeddings. On average, restricting to head entities (\texttt{head}) outperforms unrestricted selection (\texttt{all}) and consistently surpasses tail-restricted facts (\texttt{tail}), which show weaker performance. Notably, \texttt{head} achieves the best performance in most settings, though benefits vary by dataset and model. For instance, with explanation size 1, \texttt{head} outperforms \texttt{all} in TransE with WN18RR (\texttt{head: 0.598} vs. \texttt{all: 0.406}), reflecting dataset and model-specific biases.



Dataset characteristics significantly influence the effectiveness of fact restrictions. For FB15k and FB15k-237, \texttt{all} generally outperforms \texttt{head}, while for WN18 and WN18RR, restricting to head-related facts (\texttt{head}) notably improves performance. Restricting to tail-related facts (\texttt{tail}) consistently weakens performance across all datasets, with significant drops in FB15k-237 (-50\%) and WN18RR (-40\%) compared to \texttt{head}. In FB15k and WN18, where inverse relations are not removed, \texttt{tail} shows only a 9\% decline compared to \texttt{head}. This is because FB15k and FB15k-237 are dense graphs, encouraging models to balance head and tail entity modelling. In sparser datasets like WN18 and WN18RR, head entities are more significant, often representing central concepts (e.g., "person" or "organization"), while tail entities serve as hubs (e.g., "male," "New York," or "CEO") with numerous connections. Removing facts associated with hub entities has a limited impact on prediction metrics due to their less central role.


Different models exhibit varying sensitivities to fact restrictions. Restricting to tail-related facts leads to average performance drops of 8.2\%, 15\%, and 41\% for ComplEx, ConvE, and TransE, respectively, compared to \texttt{head}. TransE, with its translational operation, strongly relies on head entity and relation embeddings, making it particularly sensitive to head-related contexts and highly biased. ConvE shares similar biases but to a lesser extent, while ComplEx models symmetrical interactions between head and tail, achieving a more balanced performance. However, even ComplEx shows an 8.2\% drop, suggesting that dataset characteristics, rather than model design alone, play a significant role in determining fact preferences.

Fact restrictions should align with dataset-specific characteristics to ensure focused and meaningful explanations. Tailoring restrictions based on graph density offers a practical heuristic. Using the fact/entity ratio as a measure of density, we applied head fact restrictions for low-density datasets (ratio $<$ 10, i.e., WN18 and WN18RR) but used unrestricted selection for high-density datasets (ratio $>$ 10, i.e., FB15k and FB15k-237) to balance performance and explanation richness. While Kelpie inherently restricts explanations to head-related facts, such constraints may limit diversity and the semantic richness of explanations. Our findings emphasize the importance of flexibility in explanation strategies, enabling them to adapt to the unique properties of datasets and models.

\subsection{Ablation Study}

To assess the effectiveness of the components in our approach, ablation experiments were conducted by sequentially removing one type of rule at a time for fact scoring. This allowed us to analyze the individual contributions of the two scoring rules—CP and PT—used in our method. Table~\ref{tab:ablation} presents the results for explanation sizes 1 and 4, where \texttt{eXpath} represents the complete method using both CP and PT rules, \texttt{eXpath(w/o CP)} indicates the method without CP rules, \texttt{eXpath(w/o PT)} excludes PT rules. The results show that removing either type of rule leads to performance drops, with reductions of 27\% and 13.5\% for CP rules and 4.5\% and 6\% for PT rules, respectively. These findings underscore the critical role of CP rules in link prediction, serving as the primary mechanism for addressing complex relational patterns. While less impactful, PT rules significantly complement CP rules by improving the diversity and reliability of the explanations. 

\begin{table*}[htbp]
\caption{Comparison of explanations generated by three competent methods for representative examples. Each cell contains the $\delta \text{MRR}$ in the first row, followed by the explanation sets generated by each model.}
\vspace{-0.1cm}
\centering
\begin{tabular}{p{2.5cm} | p{2.5cm} | p{5.5cm} | p{6cm}}
\toprule
              Prediction &                                     KGEAttack &                                            Kelpie &                                          eXpath \\
\midrule
     $e_2$, award\_nominee, $e_1$ \newline(from \texttt{complex FB15k}) & [0.89]\newline $e_1$, award, $e_2$ & [L1: 0.25/L4: 0.38]\newline $e_2$, award\_nominee, Anna\_Paquin\newline $e_2$, award\_nominee, Shohreh\_Aghdashloo\newline $e_2$, award\_nominee, Julia\_Ormond\newline $e_2$, award\_nominee, Amanda\_Plummer & [L1: 0.89/L4: 0.95]\newline $e_1$, award, $e_2$\newline $e_2$, award\_nominee, Joan\_Allen\newline Tony\_Award.., award\_nominee, $e_1$\newline Academy\_Award.., award\_nominee, $e_1$ \\
     \midrule
    Porco\_Rosso, country, Japan \newline(from \texttt{conve FB15k}) &        [0.00]\newline Anime, films\_in\_this\_genre, Porco\_Rosso &       [L1: 0.62/L4: 0.74]\newline Hayao\_Miyazaki, film, Porco\_Rosso\newline Porco\_Rosso, language, Japanese\_Language & [L1: 0.73/L4: 0.84]\newline Porco\_Rosso, language, Japanese\_Language\newline Hayao\_Miyazaki, film, Porco\_Rosso\newline Fantasy, titles, Porco\_Rosso\newline Porco\_Rosso, written\_by, Hayao\_Miyazaki \\
    \midrule
    $e_3$, actor, Jonathan\_Pryce \newline(from \texttt{complex FB15k}) & [0.00]\newline $e_3$, prequel, $e_4$ & [L1: 0.00/L4: 0.58]\newline $e_4$, sequel, $e_3$\newline Keith\_Richards, film, $e_3$\newline $e_3$, actor, Keith\_Richards\newline Action\_Film, films\_in\_this\_genre, $e_3$ & [L1: 0.33/L4: 1.00]\newline $e_5$, actor, Jonathan\_Pryce\newline Jonathan\_Pryce, film, $e_5$\newline Jonathan\_Pryce, film, $e_4$\newline $e_3$, actor, Johnny\_Depp \\
\bottomrule
\end{tabular}
\label{tab:example}
\end{table*}

The impact of rule removal varies across datasets. For FB15k, CP rules prove essential, with an average performance drop of 38\%, while PT rules have less impact, suggesting that CP rules alone are sufficient to support most predictions in this dataset. On the other hand, in FB15k-237, PT rules have the greater influence, with an average drop of 42\%, whereas CP rules contribute less significantly. This discrepancy suggests that the denser and more diverse relational structures in FB15k-237 benefit from PT rules. For WN18, CP rules show a significant effect, with an average performance drop of 23\%, reflecting the importance of capturing linguistic biases through CP rules in this dataset. Interestingly, for WN18RR, neither CP nor PT rules individually cause significant performance degradation. This observation indicates that CP and PT rules are complementary, often providing overlapping support, especially in sparse datasets like WN18RR.

These results provide several key insights into the role of CP and PT rules. CP rules are foundational for addressing linguistic biases and supporting most link predictions. Even without PT rules, as in \texttt{eXpath(w/o PT)}, the method can still recommend effective explanations, emphasizing the centrality of CP rules. Meanwhile, PT rules serve as valuable complements, particularly in datasets with complex relational structures like FB15k-237, where their absence significantly impacts performance. Furthermore, the complementary nature of CP and PT rules ensures robust performance, particularly in datasets like WN18RR. These findings demonstrate that while CP rules form the backbone of effective explanations, PT rules enhance the overall credibility and diversity of explanations, particularly in datasets with diverse or dense relational structures.

\subsection{Case Study}

\begin{figure*}[t]
\includegraphics[width=\textwidth, clip, trim=40pt 20pt 20pt 20pt]{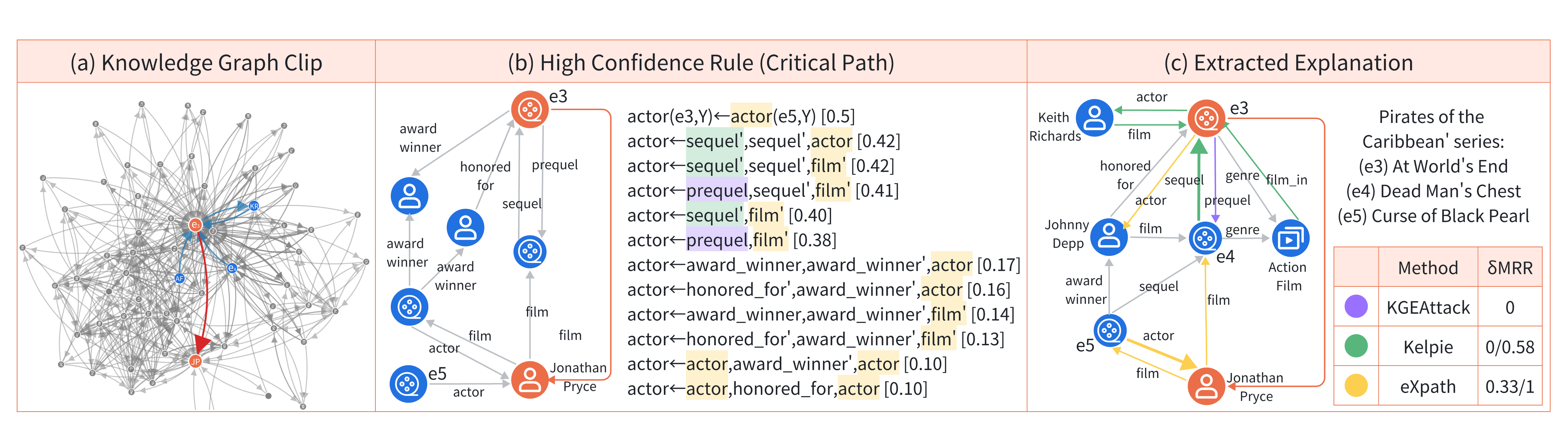}
\caption{Explanation of the fact $\langle$$e_3$, actor, Jonathan\_Pryce$\rangle$ predicted by LP models (ComplEx); (a) all 3-hop paths from head entity to tail entity. (b) Twelve high-confidence rules with $SC \geq 0.1$ identified by eXpath; (c) comparison of the explanation provided by KGEAttack (in purple edge), Kelpie (in green edges), and eXpath (in yellow edges). }
\label{example}
\end{figure*}

Table~\ref{tab:example} presents three representative cases, comparing the explanations generated by three methods: KGEAttack, Kelpie, and eXpath. Both Kelpie and eXpath can generate single-fact explanations (L1) and multi-fact explanations (L4). For clarity, some entities are represented by abbreviations due to their lengthy names: 
$e_1$ to "Frances McDormand,
$e_2$ to "Primetime Emmy Award for Outstanding Supporting Actress",
$e_3, e_4, e_5$ to the Pirates of the Caribbean series,
$e_3$ to "At World's End",
$e_4$ to "Dead Man's Chest", and
$e_5$ to "The Curse of the Black Pearl".

In the first example, the strengths of KGEAttack and eXpath are highlighted, as both methods produce highly effective explanations of the form $\langle e_1, award, e_2\rangle$, leading to a significant drop in head/tail ranks from $1/1$ to $6/106$. Both systems support this explanation with a compelling CP rule: \texttt{award\_nominee $\leftarrow$ award' [0.815]}, which intuitively links \texttt{award\_nominee} and \texttt{award} as inverse relations (number in the square bracket represents standard confidence (SC) of the rule). In contrast, Kelpie produces weaker results despite using four facts. We observe that Kelpie's explanations are based on facts such as $\langle e_2, award\_nominee, X\rangle$; however, without supporting ontological rules, it is difficult to justify the adequacy of these explanations, underscoring Kelpie's limitations when compared to rule-based systems like eXpath.

The second example reverses the trend, with Kelpie (L1) outperforming KGEAttack. KGEAttack generates an intuitive PT rule: \texttt{country(X, Japan) $\leftarrow$ films\_in\_this\_genre(Anime, X) [0.846]}, suggesting that a film in the Anime genre is likely associated with Japan. eXpath surpasses both methods by providing a more comprehensive explanation, combining multiple rules:

\begin{itemize}
    \item \texttt{country(X, Japan) $\leftarrow$ language(X, Japanese) [0.669]}
    \item \texttt{country $\leftarrow$ language, language', country [0.311]}
    \item \texttt{country $\leftarrow$ language, language', nationality [0.194]}
    \item \texttt{country $\leftarrow$ language, titles, country [0.122]}
\end{itemize}

All four rules identified by eXpath are high-confidence ($SC \geq 0.1$), with their standard confidence (SC) values indicated in brackets. While the SC of each rule is lower than that of KGEAttack's rule, collectively, they yield a cumulative confidence greater than $0.9$. This demonstrates that relying solely on simple or individual rules, as KGEAttack does, risks overlooking valuable data signals. Kelpie's explanation shares two facts with eXpath's initial rules but is heavily based on empirical signals from the embedding model and lacks the clarity and reliability of rule-based approaches.

The third example involves the prediction $\langle e_3$, actor, Jonathan Pryce$\rangle$. Kelpie and eXpath offer four-fact explanations, with single-fact versions highlighted in bold. Notably, eXpath (L4) delivers the most effective explanation, achieving near-perfect attack effectiveness ($\delta \text{MRR}$ approaching $1$), while Kelpie (L4) also performs well ($\delta \text{MRR} = 0.58$). In contrast, explanations from KGEAttack and Kelpie (L1) are largely ineffective. The consistent performance of multi-fact explanations highlights the importance of combining multiple facts, especially in dense datasets like \texttt{FB15k} and \texttt{FB15k-237}, where removing a single fact often fails to impact the prediction.

Kelpie provides fact-based explanations but fails to justify the relevance of these facts in supporting the prediction. One fact, $\langle e_4, sequel, e_3\rangle$, is supported by three high-confidence rules, including \texttt{actor $\leftarrow$ sequel', film' [SC=0.40]}, while the remaining facts lack direct relevance. Removing this fact leaves the reverse relation $\langle e_3, prequel, e_4\rangle$, which still supports the prediction, undermining the explanation's validity. KGEAttack also proposes a single attacking fact, $\langle e_3, prequel, e_4\rangle$, supported by the rule \texttt{actor $\leftarrow$ prequel, film' [SC=0.38]}. Although intuitive, this 2-hop CP rule fails for the same reason as Kelpie: the reverse relation maintains the prediction, rendering the explanation insufficient.

In contrast, eXpath provides path-based explanations, combining selected facts with supporting rules. For example, the highest-scoring fact, $\langle e_5, actor, Jonathan\_Pryce\rangle$, is supported by one PT and five CP rules, as detailed in Figure~\ref{example}(b). These rules collectively contribute to a cumulative score exceeding $0.9$. Unlike KGEAttack, which focuses only on 2-hop CP rules, eXpath incorporates longer, more complex rules, capturing additional data signals. As shown in Figure~\ref{example}(b), eXpath's four facts comprehensively cover all critical paths from $e_3$ to Jonathan Pryce, yielding a nearly perfect explanation for the prediction.

An interesting observation is that most facts selected by eXpath relate to the tail entity rather than the head entity (shown in Figure~\ref{example}(c)). As depicted in Figure~\ref{example}(a), the head entity ($e_3$) is associated with 96 triples. In contrast, the tail entity (Jonathan Pryce) is connected to only 32, making tail relations sparser and more critical for prediction. By prioritizing tail-related facts, eXpath produces more effective explanations. In contrast, Kelpie relies predominantly on head entity features, often getting trapped in local optima and missing broader contextual signals. Meanwhile, KGEAttack selects rules randomly from those it satisfies, leading to highly varied explanations and limited reliability.

These case studies demonstrate the superior performance of eXpath in generating semantically rich and effective explanations. By leveraging comprehensive rule-based reasoning and integrating multiple facts, eXpath strikes an optimal balance between interpretability and explanatory power, consistently outperforming alternative methods.